\documentclass[nonacm]{acmart}
\usepackage{amsmath}

\newcommand{\KLD}[2]{D_{\mathrm{KL}} (  #1 \left| \right| #2 )}

\setcopyright{none}

\begin{document}

\title{Accelerated Design and Deployment of Low-Carbon Concrete for Data Centers}

\author{Xiou Ge}
\authornote{This work was performed when the author was at the University of Illinois Urbana-Champaign.}
\affiliation{%
  \institution{Ming Hsieh Department of Electrical and Computer Engineering, University of Southern California}
  \country{USA}
}

\author{Richard T. Goodwin}
\affiliation{%
  \institution{IBM Thomas J.\ Watson Research Center}
  \country{USA}
}

\author{Haizi Yu}
\affiliation{%
  \institution{Knowledge Lab, University of Chicago}
  \country{USA}
}

\author{Pablo Romero}
\affiliation{%
 \institution{Department of Civil and Environmental Engineering, University of Illinois Urbana-Champaign}
 \country{USA}
}

\author{Omar Abdelrahman}
\affiliation{%
 \institution{Department of Civil and Environmental Engineering, University of Illinois Urbana-Champaign}
 \country{USA}
}

\author{Amruta Sudhalkar}
\affiliation{%
 \institution{Meta}
 \country{USA}
}

\author{Julius Kusuma}
\affiliation{%
 \institution{Meta}
 \country{USA}
}

\author{Ryan Cialdella}
\affiliation{%
 \institution{Ozinga Ready Mix}
 \country{USA}
}

\author{Nishant Garg}
\affiliation{%
 \institution{Department of Civil and Environmental Engineering, University of Illinois Urbana-Champaign}
 \country{USA}
}

\author{Lav R. Varshney}
\affiliation{%
  \institution{Department of Electrical and Computer Engineering, University of Illinois Urbana-Champaign}
  \country{USA}
}

\renewcommand{\shortauthors}{Ge, et al.}

\begin{abstract}
Concrete is the most widely used engineered material in the world with more than 10 billion tons produced annually. Unfortunately, with that scale comes a significant burden in terms of energy, water, and release of greenhouse gases and other pollutants; indeed $8\%$ of worldwide carbon emissions are attributed to the production of cement, a key ingredient in concrete. As such, there is interest in creating concrete formulas that minimize this environmental burden, while satisfying engineering performance requirements including compressive strength.  Specifically for computing, concrete is a major ingredient in the construction of data centers.

In this work, we use conditional variational autoencoders (CVAEs), a type of semi-supervised generative artificial intelligence (AI) model, to discover concrete formulas with desired properties. Our model is trained just using a small open dataset from the UCI Machine Learning Repository joined with environmental impact data from standard lifecycle analysis. Computational predictions demonstrate CVAEs can design concrete formulas with much lower carbon requirements than existing formulations while meeting design requirements. Next we report laboratory-based compressive strength  experiments for five AI-generated formulations, which demonstrate that the formulations exceed design requirements.  The resulting formulations were then used by Ozinga Ready Mix -- a concrete supplier -- to generate field-ready concrete formulations, based on local conditions and their expertise in concrete design.
Finally, we report on how these formulations were used in the construction of buildings and structures in a Meta data center in DeKalb, IL, USA.  Results from field experiments as part of this real-world deployment corroborate the efficacy of AI-generated low-carbon concrete mixes.
\end{abstract}




\begin{teaserfigure}
  \includegraphics[width=\textwidth]{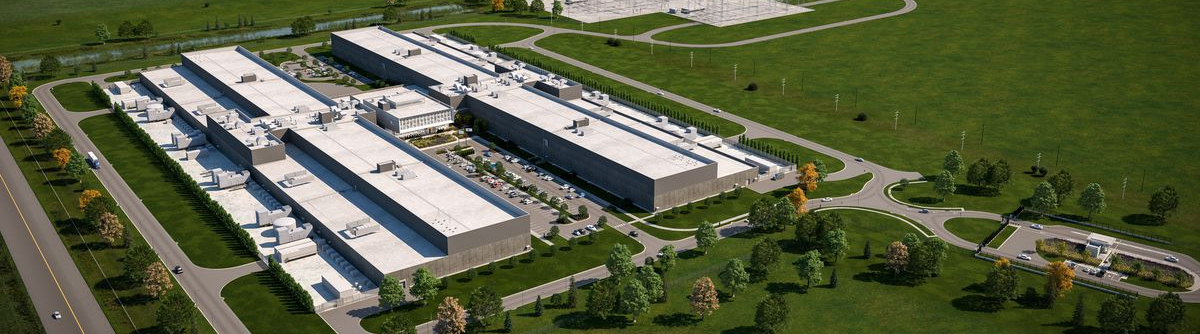}
  \caption{Depiction of Meta's data center in DeKalb, IL, USA where low-carbon concrete discovered by artificial intelligence methods has been tested.}
  \label{fig:teaser}
\end{teaserfigure}

\maketitle

\section{Introduction}

Is it possible to formulate concrete mixtures with standard ingredients so that they have half as much embodied carbon as traditional formulations, yet are just as strong?  We answer this question affirmatively, through generative artificial intelligence (AI) algorithms, lab testing, and field trials at Meta's data center in DeKalb, IL, USA. 

\begin{figure}
\centering
\includegraphics[width=\textwidth]{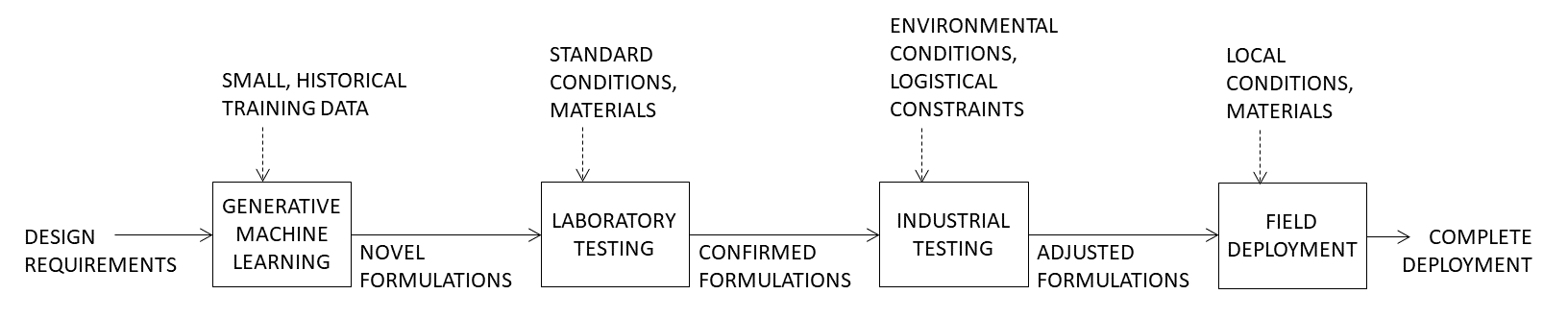}
\caption{Schematic of the accelerated design and deployment process we undertook, going from training a generative algorithm to full deployment in the field.}
\label{fig:blockdiagram}
\end{figure}

The building sector accounts for a significant fraction of overall energy consumption and pollution worldwide, including residential, commercial, and infrastructural structures such as the construction of computing infrastructure like data centers. Concrete -- including its key ingredient, cement -- is one of the most energy-intensive and polluting building materials to fabricate, but is also the most widely used engineered material in the world with more than 10 billion tons produced annually. Unfortunately, with that scale comes a significant burden in terms of energy, water, and release of greenhouse gases and other pollutants; indeed $8\%$ of worldwide carbon emissions are attributed to the production of cement \cite{Nature2021, EllisBCPC2020}.
As part of larger efforts to use machine learning to address problems in climate change \cite{Rolnick2019}, here we  approach the design of low-carbon concrete using generative machine learning (ML) algorithms.  But we do not just stop at the algorithm stage; rather we further demonstrate the efficacy of our approach via laboratory experiments, translation to industrial practice, and field tests in a real data center location, that confirms the practicality of our approach. Figure \ref{fig:blockdiagram} depicts how we move from stage to stage of our research.

Although concrete has been designed, developed, and optimized for more than 7000 years across numerous civilizations, yielding famous structures such as the Pantheon in Rome, the Willis Tower in Chicago, and the Burj Khalifa in Dubai \cite{Palley2010, JahrenS2018, Courland2011}, there are still numerous open questions regarding its properties. As a material, concrete is primarily composed of water, fine/coarse aggregate, cement, and other cementitious materials such as slag and fly ash, which are industrial byproducts. Only very recent work in computational statistical mechanics and theory has started to reveal how cement cohesion happens at the nanoscale level to give concrete its strength over time \cite{GoyalPIUDPED2021}, but physics-based models are unable to accurately predict mechanical properties of concrete.  As such, there is a growing scientific literature on using supervised machine learning to predict concrete strength based on its composition  
\cite{Yeh1998, YoungHPGS2019,ChaabeneFN2020}.   Most such extant ML predictors draw on ensemble methods such as random forests, which have limited scientific interpretability \cite{AriaCG2021} and so it is unclear how to directly use insights from predictive algorithms to design novel concrete formulations with particular desired properties.  

Since the first presentation of our work in 2019 on generating novel concrete formulations  \cite{GeGGKMV2019} -- which was restricted to just the algorithmic stage -- there have been several papers that put multi-objective optimization procedures around predictive ML for design of concrete formulations. All resort to meta-heuristics for global optimization \cite{DeRousseauKS2018,DabbaghiTNNDY2021,ZhangHMN2021,ZhangHWM2020,MahjoubiBGMB2021,RenZZZXCWCW2022,HanSKHK2020}, an approach that is not only very inefficient computationally, but may also use predictive ML algorithms well-outside the domain of their training data where their accuracy is quite limited \cite{KshetryV2019, MeinkeH2020}.  These recent works do not perform laboratory or field experiments to validate the results of the algorithms.  We take an alternative approach and develop \emph{generative} ML models rather than predictive ones: such models directly generate new concrete formulations rather than requiring an outer optimization loop and make as best use of training data as they can.  

Indeed, recent advances in machine learning and artificial intelligence have enabled machines to generate very high-quality artifacts, such as images of realistic looking faces with certain desired properties \cite{YanYSL2016}, or natural language with certain desired topics and styles \cite{KeskarMVXS2019}. In this work, we use an extension of the popular generative modeling framework of variational autoencoders (VAEs) \cite{KingmaW2019} that allows control of attributes.  This extension is conditional variational autoencoders (CVAEs) \cite{SohnLY2015}, a type of semi-supervised generative model \cite{KingmaRMW2014}, which we can use to generate concrete formulas with desired properties. 
Note that although deep generative models have been applied in materials and molecules discovery---\cite{Gomez-Bombarelli2018} uses VAEs based on recurrent neural networks for chemical design in which molecules are encoded as strings and \cite{RampasekHSHG2019} uses VAEs to improve the accuracy of drug response predictions---they have not previously been used in the context of aggregate materials. In fact, the field of ML-based accelerated materials discovery is oriented away from aggregate materials such as concrete \cite{LookmanEAB2018}, e.g.\ the Materials Genome Initiative \citep{Jain2013} is focused on inorganic compounds, nonporous materials, and electrodes.

We demonstrate CVAEs can design concrete formulas with lower emissions and natural resource usage while meeting design requirements, computationally and also in laboratory and field experiments. In the computational phase before experimental testing, we also train regression models to predict the environmental impacts and strength of generated formulas. This provides initial insight to civil engineers in creating formulas that meet structural needs and best addresses local environmental concerns.  

To assess environmental impacts, the Cement Sustainability Initiative (CSI) developed the Environmental Product Declaration (EPD) tool to facilitate the generation of sector-specific EPDs for cement, concrete, clinker, lime, and plaster.  EPD is a voluntary declaration that provides quantitative information about the environmental impact of a product, using life-cycle assessment (LCA) methodology and verified by an independent third party. The cloud-based tool was designed to be easy-to-use, to facilitate the process overall, and to reduce the costs of preparing cement and concrete EPDs. In this work, we join this with open, historical concrete formulation and strength data from the UCI Machine Learning Repository \cite{Yeh1998}.

\subsection{Sustainable Data Centers}
Much of the conversation on sustainable computing infrastructure, such as data centers, focuses on operational sustainability \cite{MasanetSLSK2020, Gao2014} by ensuring facilities use as little power as possible, use renewable energy sources, and have minimal impacts on local areas via co-generation and related techniques.  Yet, besides communication, computation, and storage of digital information that operational efficiency captures, there is also the expressly material infrastructure in buildings and structures \cite{Furlong2021}.  This is also starting to get attention \cite{KrieghMS2021}, with a particular focus on embodied carbon in building infrastructure such as concrete.  Embodied carbon includes emissions caused by extraction, manufacture, transportation, assembly, maintenance, replacement, deconstruction, disposal, and end of life aspects of the materials and systems that make up a building.  As such, we conduct our field deployment in the context of a new data center being built in DeKalb, IL, USA, see Figure~\ref{fig:teaser}.

\subsection{Paper Organization}
The remainder of the paper is organized as follows. We first describe the data set and the CVAE model details. Next we give results, first showing the average percentage reduction environmental impact achieved by generated better-performing concrete formulas. We then show strength spectrum plots in the 3D environmental impact space which could be turned into a visualization tool for concrete designers. Third, we evaluate the performance of strength conditioned generation of the trained model. After detailing the computational phase of our research, we then present results from laboratory experiments and from full field deployments.

\section{Generative Machine Learning}

\begin{figure}
\centering
\includegraphics[width=4in]{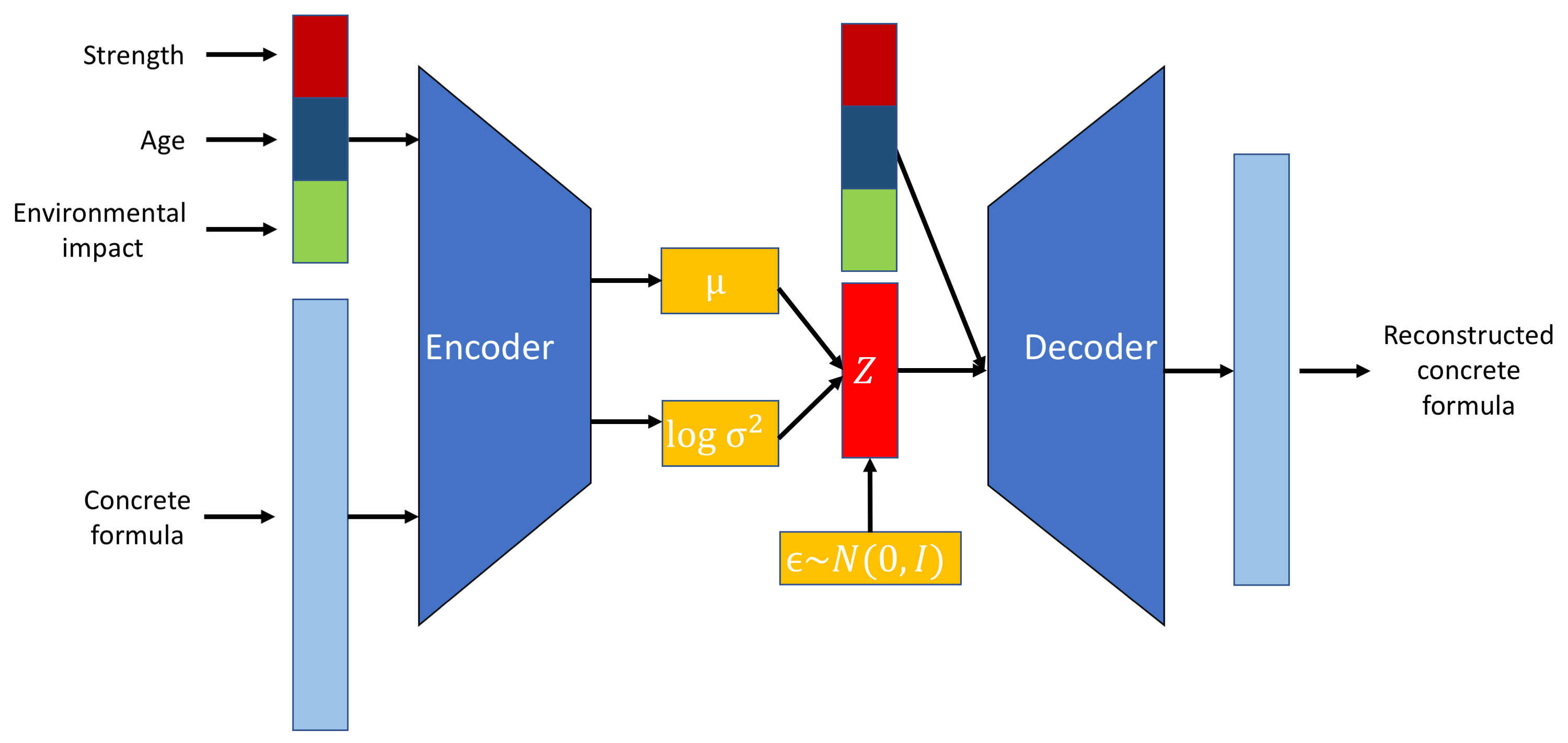}
\caption{CVAE model structure}
\label{fig:CVAE_structure}
\end{figure}

\subsection{Training Data}
We train our model using the Concrete Compressive Strength Data Set \citep{Yeh1998} openly available from the UCI Machine Learning Repository. It has 1,030 training examples, with seven continuous features describing the amount of constituent material such as cement, aggregates, and water. Compressive strength, after a particular curing time (age), of each concrete formula is also given. In addition, we use the CSI EPD tool to estimate the environmental impact of each concrete formula. The EPD tool produces 12 continuous features characterizing the concrete environmental impact. Among these, we largely focus on global warming potential (GWP) measured using embodied CO$_2$, acidification potential (AP) measured using embodied SO$_2$, and concrete batching water (CBW) consumption.  Future work could more explicitly focus on the other nine environmental impacts.

\subsection{Generative Model}
Our model is based on a variant of the VAE \cite{KingmaW2019} called CVAE \cite{SohnLY2015} as shown in Figure~\ref{fig:CVAE_structure}. Like other generative models, the goal is to estimate the data distribution $p(y)$ and to generate realistic new samples from that distribution \citep{Doersch2016}. What makes CVAE different from VAE is that instead of merely generating realistic samples from the data distribution $p(y)$ randomly, we generate from the conditional distribution $p(y|x)$ which give us control over the underlying properties of generated data by conditioning on different values of $x$. 

We interpret the variables in the conditional generative model as follows: $x$ represents the side information of a formula including the strength, age, and environmental impacts, $y$ represents the constituent material amount of a formula, and $z$ represents the latent variables. Like the VAE, a CVAE consists of an encoder $q_\phi(z|x,y)$ that maps the data points to latent codes and a decoder $p_\theta(y|x,z)$ that reconstructs the data points from latent codes. The decoder and encoder are implemented as neural networks where $\phi$ and $\theta$ are the respective network parameters. Since the goal is to generate realistic concrete samples with desired properties, we want to maximize the log-likelihood of the data distribution model $\log{p_\theta(y^{(i)}|x^{(i)})}$. Since the data  distribution $p_\theta(y|x)$ and the posterior distribution $p_\theta(z|x,y)$ are both intractable, we maximize the Evidence Lower Bound (ELBO), $\mathcal{L}$, instead. The loss function of CVAE is therefore:
\begin{equation}
	\log{p_\theta(y^{(i)}|x^{(i)})} \geq \mathbf{E}_{z\sim q_\phi(z|x,y)}[\log{p_\theta(y^{(i)}|z,x^{(i)})}]
	- \KLD{q_\phi(z|x^{(i)},y^{(i)})}{p_\theta(z|x^{(i)})} = \mathcal{L} \mbox{.}
\end{equation}

\subsection{Implementation Details}

In our model, the encoder network consists of four fully-connected layers with 25 neurons on the first layer, 20 neurons on the second layer, followed by two parallel layers with two neurons on each which represent the mean and log variance respectively. The prior is set to be an isotropic Gaussian distribution with zero mean and unit variance $p(z)=\mathcal{N}(0,\mathcal{I})$. The reparameterization trick is performed to make the sampling step differentiable and enable backpropagation for training. The decoder network consists of two fully-connected layers with 20 neurons on the first layer and 25 neurons on the second layer. Rectified linear unit (ReLU) activation functions are applied to all layers except the output layer of the decoder, where we use sigmoid activation since we scale our data to $[0,1]$. The model is trained end-to-end with a version of stochastic gradient descent: the Adam optimizer with learning rate of $0.001$ and batch size of $10$ \cite{KingmaB2015}.

\begin{figure}
\centering
\includegraphics[width=4in]{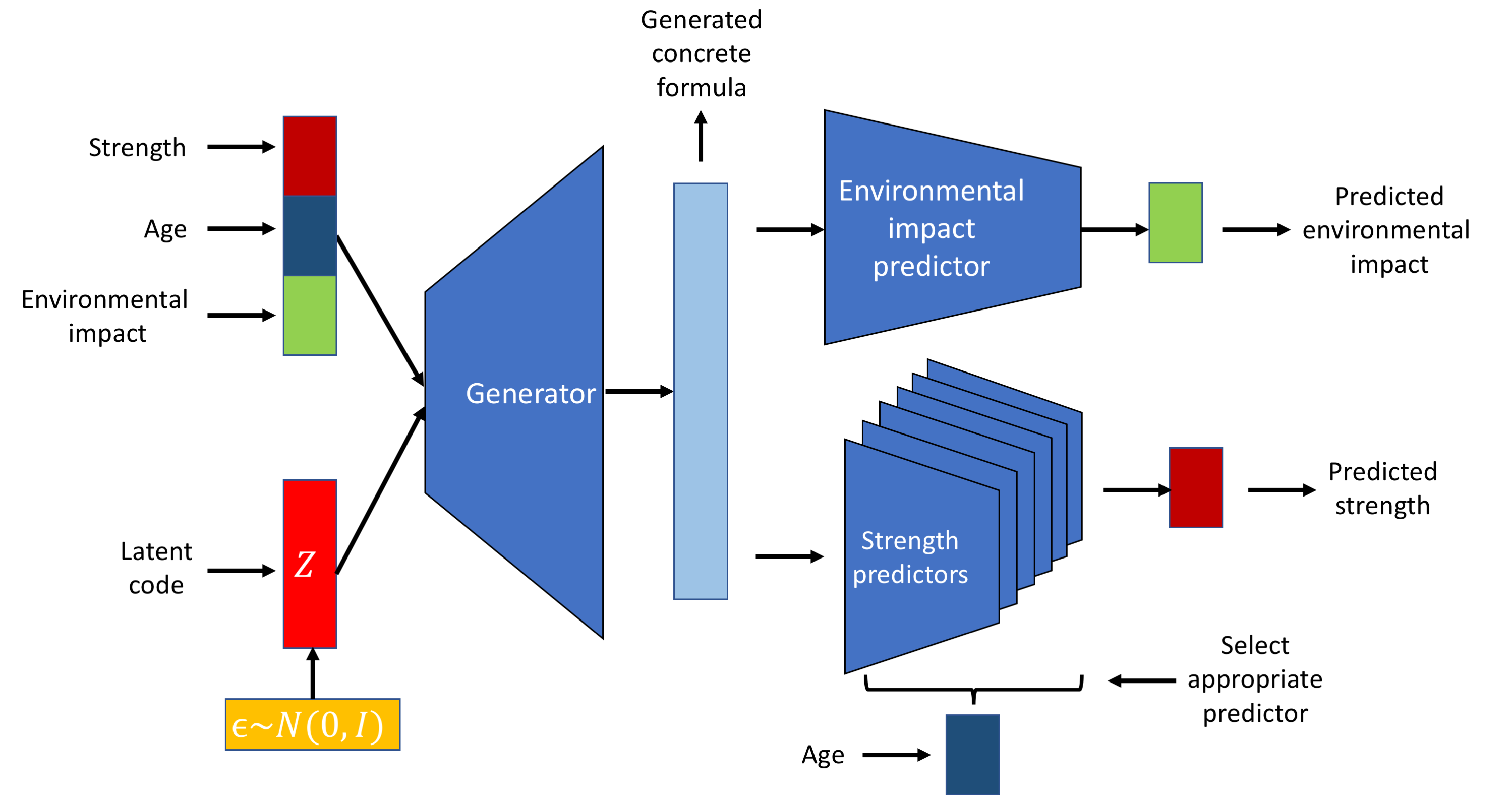}
\caption{Generating new concrete formulas and evaluating their properties}
\label{fig:generator_predictor}
\end{figure}

\subsection{Property Predictors}
We also trained neural network-based regression models as shown in Figure~\ref{fig:generator_predictor} using the dataset that we described above to predict the environmental impact and strength of concrete formulas. Since the compressive strength is dependent on the age of concrete, we trained separate compressive strength predictors for each age group. The purpose of the predictors is twofold. First, we can measure how well the properties of generated samples match the desired properties given as conditioning variables during data generation. Second, we can make fair comparisons between extant and generated concrete formulas in terms of the environmental impact. We experimented with three different types of regression models, namely linear regression, decision tree regression, and neural network regression. Although linear regression can achieve comparable performance with decision tree regression and neural network regression, it often predicts far-out-of-range values for newly generated concrete formulas. The neural network regression has slightly better performance than the decision tree regression and therefore we use the former for prediction tasks. The performance of the neural network regression models for global warming potential (GWP), acidification potential (AP), and concrete batching water (CBW) consumption are shown in Table~\ref{table:Predictor_Performance}. The performance of the strength predictors are shown in Table~\ref{table:Strength_Predictor_Performance}.
 
\begin{table}
\centering
\begin{tabular}{rrrrr}
\toprule
Metric & GWP & AP & CBW \\
 & (kg CO$_2$ eq./$m^3$) & (kg SO$_2$ eq./$m^3$) & ($m^3$) \\
\midrule
MAE & 7.187  & 0.019 & 0.003   \\
RMSE & 9.374  & 0.040 & 0.006  \\
$R^2$ & 0.979  & 0.974 & 0.881 \\
\bottomrule
\end{tabular}
\caption{Environmental Impacts Predictor Performance}
\label{table:Predictor_Performance}
\end{table}

\begin{table}
\centering
\begin{tabular}{rrrrrrr}
\toprule
& 
& \multicolumn{5}{c}{Predictor Performance (MPa)}\\
\cmidrule(r){2-7}
Age & $\leq$3 & 7 & 14 & 28 & 56 & $\geq$90 \\
\midrule
MAE & 2.985 & 3.850 & 3.378 & 6.015 & 5.093 & 4.457\\
RMSE & 0.222 & 0.201 & 0.163 & 0.227 & 0.124 & 0.125\\
$R^2$ & 0.819 & 0.870 & 0.703 & 0.679 & 0.795 & 0.789\\
\bottomrule
\end{tabular}
\caption{Strength Predictor Performance}
\label{table:Strength_Predictor_Performance}
\end{table}

\section{Computational Results}

\subsection{Concrete Formulas with Reduced Environmental Impact}

\begin{table}
  \centering
  \begin{tabular}{rrrrr}
    \toprule
     & & \multicolumn{3}{c}{Average Reduction ($\%$)}\\
    \cmidrule(r){3-5}
    Age & Strength &  GWP & AP & CBW \\
    (day) & (MPa) &  (kg CO$_2$ eq./$m^3$) &  (kg SO$_2$ eq./$m^3$)& ($m^3$) \\
    
    \midrule
    $\leq$3 & 30$\pm$1 & 0.80 & 1.83 & 5.47 \\
            & 40$\pm$1 & 7.74 & 1.59 & 0.26 \\
    \cmidrule(r){1-5}
    7 & 30$\pm$1 & 19.69 & 3.94 & 7.58 \\
      & 40$\pm$1 & 25.45 & 11.33 & 5.03 \\
    \cmidrule(r){1-5}
    14 & 20$\pm$1 & 2.20 & 5.72 & 10.64 \\
       & 60$\pm$1 & 42.45 & 21.09 & 5.17 \\
    \cmidrule(r){1-5}
    28 & 70$\pm$1 & 21.62 & 6.66 & 3.32 \\
       & 80$\pm$1 & 27.44 & 8.40 & 4.15 \\
    \cmidrule(r){1-5}
    56 & 40$\pm$1 & 4.38 & 2.95 & 7.04 \\
       & 50$\pm$1 & 14.38 & 3.23 & 3.64 \\
       & 70$\pm$1 & 30.26 & 23.75 & 1.32 \\
       & 80$\pm$1 & 5.88 & 1.33 & 3.46 \\
    \cmidrule(r){1-5}
    $\geq$90 & 80$\pm$1 & 30.58 & 6.91 & 4.11 \\
    \bottomrule
  \end{tabular}
  \caption{Average environmental impact reduction achieved by better performing generated samples}
\label{table:environmental_impact_reduction}
\end{table}

\begin{table}[t]
  \centering
  \begin{tabular}{rrr}
    \toprule
    Strength (MPa) & 30$\pm$1 & 40$\pm$1 \\
    \cmidrule(r){2-3}
    Constituent Material & \multicolumn{2}{c}{Amount (kg per $m^3$)}\\
    \midrule
    Cement & 186.4 & 259.0\\
    Blast Furnace Slag & 236.7 & 288.6\\
    Fly Ash & 107.1 & 58.8\\
    Water & 142.3 & 142.5\\
    Superplasticizer & 22.3 & 26.1\\
    Coarse Aggregate & 901.4 & 868.6\\
    Fine Aggregate & 717.2 & 763.0\\
    \bottomrule
  \end{tabular}
  \caption{Sample concrete formula with reduced environmental impact}
  \label{table:sample}
\end{table}

\begin{figure}
    \centering 
  \includegraphics[width=2.5in]{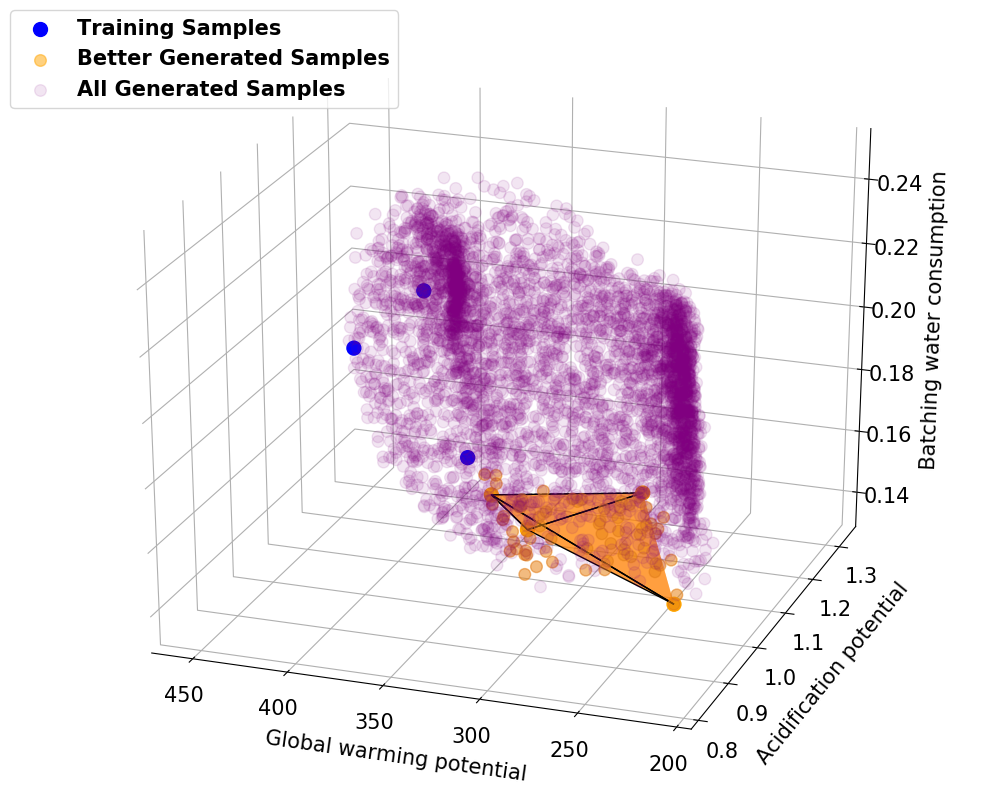} \ 
  \includegraphics[width=2.5in]{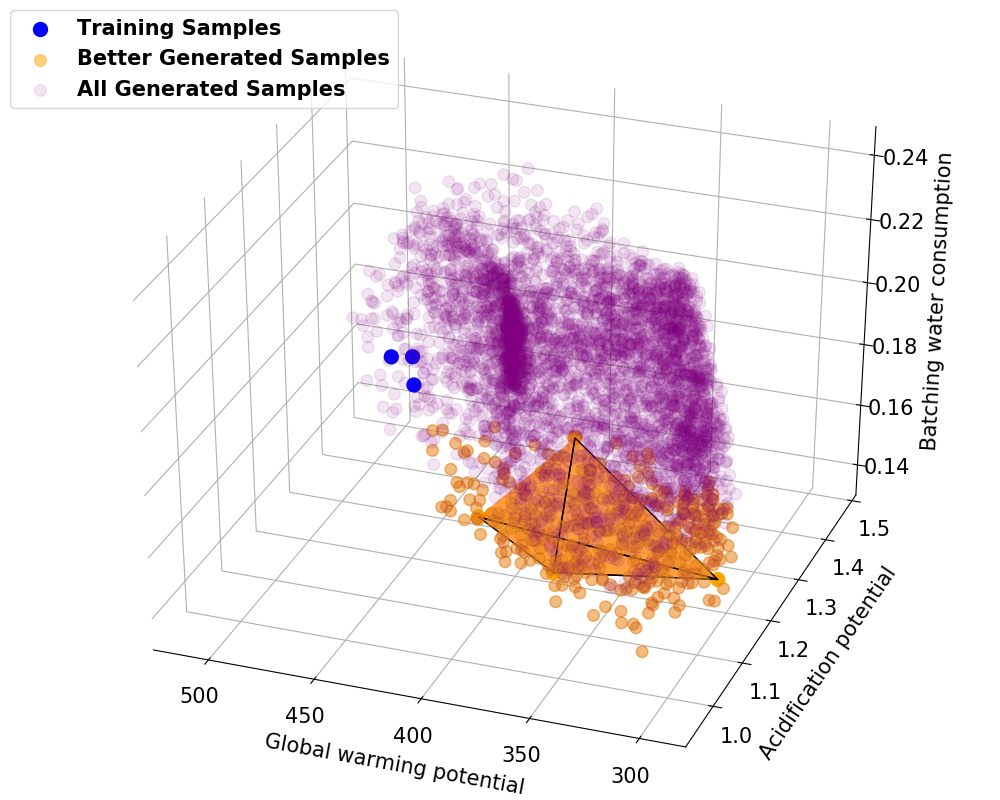} \\
    (a) \qquad\qquad\qquad\qquad\qquad\qquad\qquad\qquad\qquad\qquad (b)
\caption{Approximated hull of generated samples from archetypal analysis, training samples, and all generated samples for specific curing time and strength level. (a) Curing time = 7 days, Strength = 30$\pm$1 MPa. (b) Curing time = 7 days, Strength = 40$\pm$1 MPa.}
\label{fig:convex_hull}
\end{figure}

To demonstrate that the generative algorithm discovers new concrete formulas with reduced environmental impacts, we compared the GWP, AP, and CBW values between existing concrete formulas in the training set \cite{Yeh1998} and generated formulas with the same age and similar strength. For each concrete age group, we generate 60,000 concrete formulas. Both the strength and the environmental impact inputs to the generator are produced by randomly sampling from the standard uniform distribution whereas the latent code input is produced by randomly sampling from the standard bivariate normal distribution. We then use the trained environmental impact predictor and strength predictor for the corresponding age group to evaluate environmental impact and strength of the newly generated formulas. We count the number of generated samples having lower environmental impact than the best observed values for extant samples in all 3 dimensions. We also measured the average percentage reduction in environmental impact for the better-performing samples as compared to extant samples in the training set.  

Notice that this performance metric of \emph{conditional average improvement} is quite conservative, as it is not just considering the best generated formulations but the whole ensemble of formulations that are better than formulations in the training dataset along all three environmental performance dimensions.

Results in Table~\ref{table:environmental_impact_reduction} show significant improvements in environmental impact. Taking 14-day strength around 60 MPa, notice that the conditional average reduction for carbon (GWP) can be as high as 42\%, while also achieving conditional reduction for sulfur (AP) as high as 21\%.  As noted, conditional average improvement/reduction are very conservative performance metrics.  When we look at some of the best specific formulations that emerge from the generative algorithm, we will see more than 50\% reduction in carbon. An example is  given in Table~\ref{table:sample}.  

We constructed an approximated convex hull that encloses a majority of the better performing points in the three-dimensional environmental metric space as shown in Figure~\ref{fig:convex_hull}. From the diagram we can also see that there is an opportunity to trade off different types of environmental impact. In Table~\ref{table:sample}, we show one specific generated concrete formula that is nearest neighbor to one of the extremal points used to construct the convex hull, for strength of 30$\pm$1 MPa and 40$\pm$1 MPa respectively. 

\subsection{Extrapolative Generation}
\begin{figure}
\centering
\includegraphics[width=3.5in]{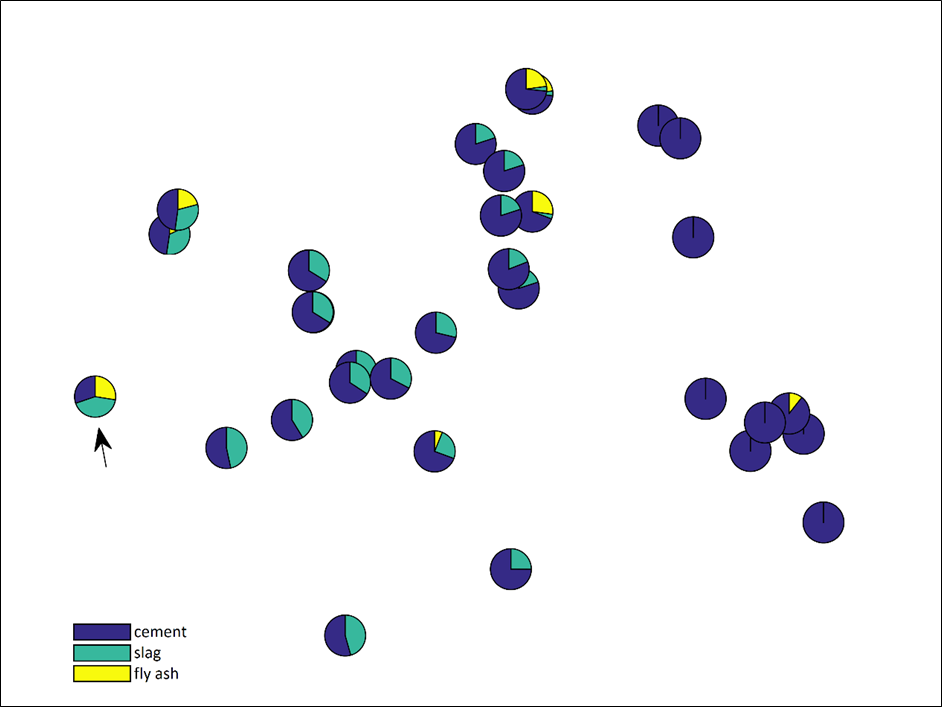}
\caption{Isomap embedding of a novel AI-generated concrete formulation (marked by arrow), together with 38 formulations from the training data having similar 28-day strength.  Placement is governed by all 7 ingredient dimensions but pie chart markers only show the cementitious ingredients.  Note that isomap dimensions are unsupervised and must be interpreted in post hoc manner.}
\label{fig:isomap}
\end{figure}

A key hallmark of creativity is novelty: the ability to generate ideas or artifacts that are beyond the training dataset.  Given the concrete formulations in the training data do not meet our desired sustainability performance metrics, we need creativity. We aim for generative algorithms that are extrapolative rather than simply interpolative in the training data \cite{DasV2022}.  To see that our generated formulations are indeed novel, consider a visualization based on the isomap dimensionality reduction technique \cite{TenenbaumSL2000} in Figure~\ref{fig:isomap} for Mix 4 from Table~\ref{tab:mix_composition} (chosen arbitrarily) together with 38 mixes from the training data \cite{Yeh1998} that have similar 28-day compressive strength.  The isomap spatial layout aims to preserve distances among points in the high-dimensional space as closely as possible in the low-dimensional space and is based on all ingredients \{cement, slag, fly ash, water, superplasticizer, coarse aggregate, and fine aggregate\}, whereas the embedded ``pie charts as markers'' only show the fractions of the first three ingredients \{cement, slag, fly ash\} for ease of understanding.

As can be observed, whether considering only the first three ingredients, or considering all of the ingredients, the new formulation (marked by an arrow) is distinct from the others and therefore novel.  In this sense, this new formulation is pushing the boundary of human creativity.  The design principle espoused in the AI-generated formulation is to considerably decrease cement by replacing with other cementitious materials such as fly ash and slag.

\subsection{Visualization for Concrete Design}

\begin{figure}
    \centering 
  \includegraphics[width=1.8in]{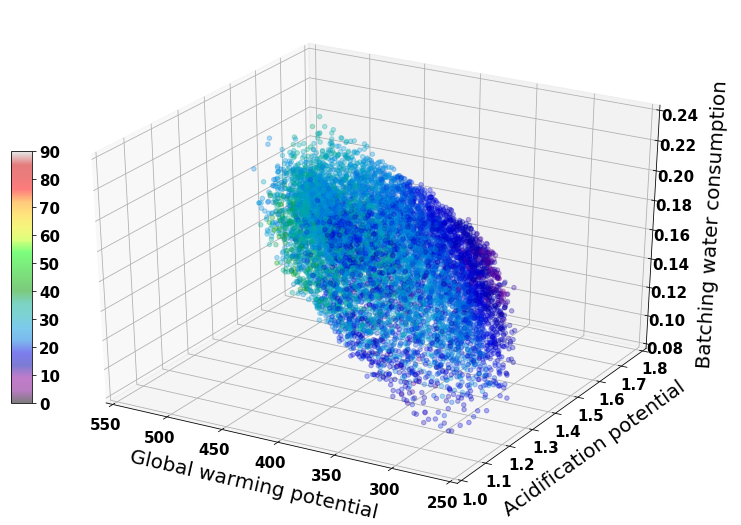}
  \includegraphics[width=1.8in]{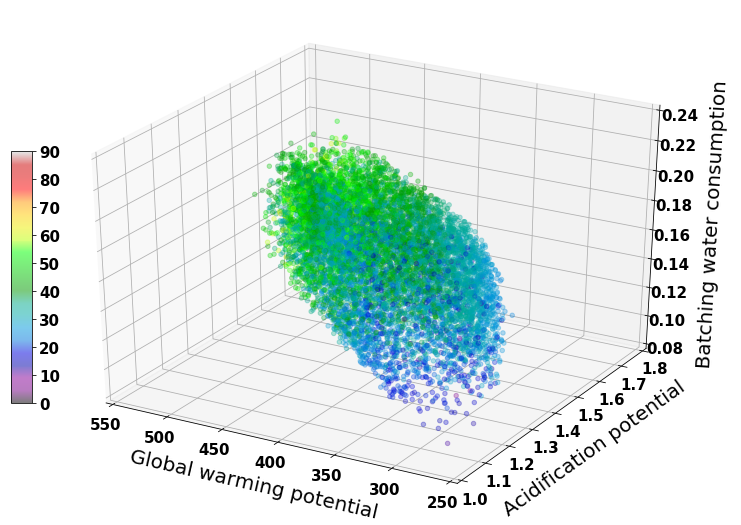}
  \includegraphics[width=1.8in]{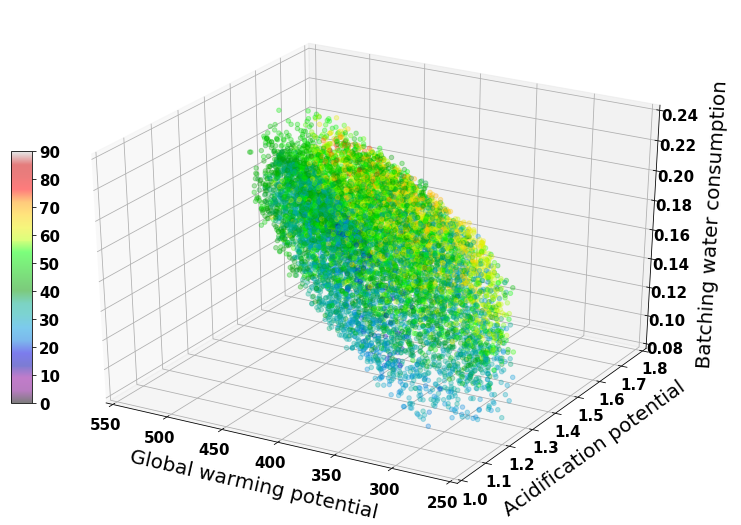} \\

(a) \qquad\qquad\qquad\qquad\qquad\qquad\qquad (b) \qquad\qquad\qquad\qquad\qquad\qquad\qquad (c) \\ 

\includegraphics[width=1.8in]{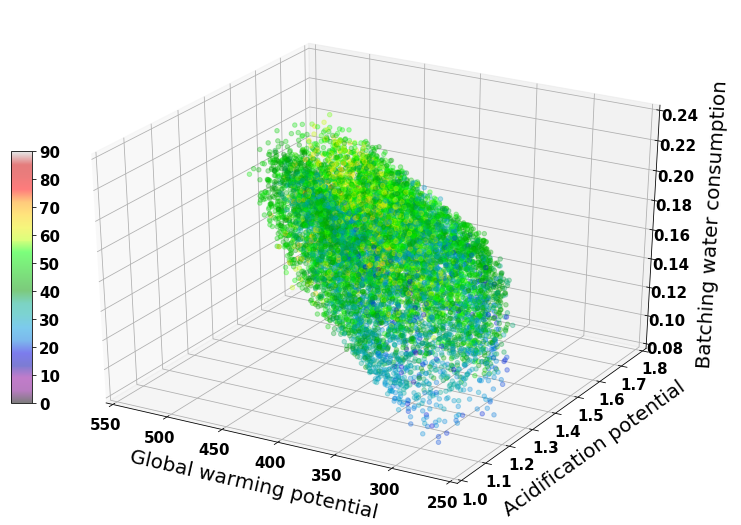}
\includegraphics[width=1.8in]{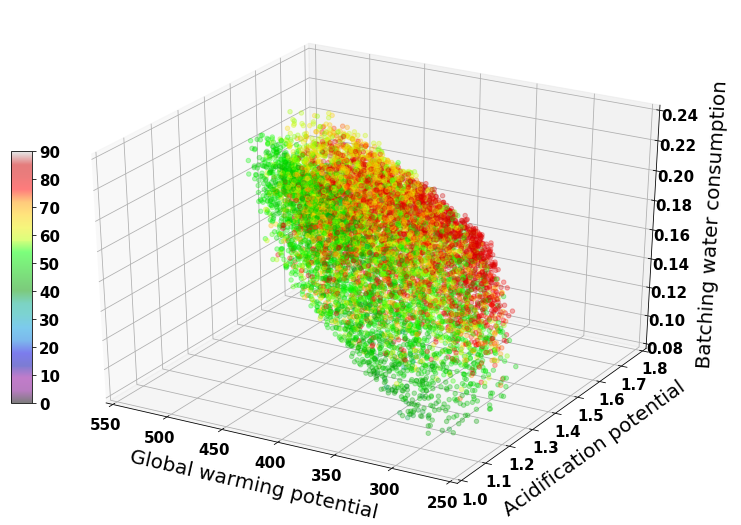}
\includegraphics[width=1.8in]{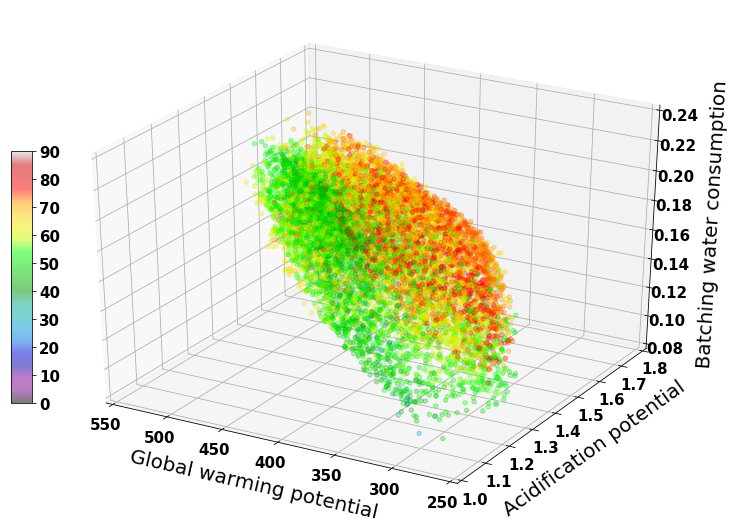} \\

(d) \qquad\qquad\qquad\qquad\qquad\qquad\qquad (e) \qquad\qquad\qquad\qquad\qquad\qquad\qquad (f)

\caption{Strength spectrum of generated concrete formulas for different concrete curing time plotted in 3D environmental impacts space, where color indicates strength. (a) $\leq$ 3 days. (b) 7 days. (c) 14 days. (d) 28 days. (e) 56 days. (f) $\geq$90 days.}
\label{fig:strength_env_profile}
\end{figure}

On top of the three-dimensional environmental impact design space that we mentioned earlier, we also color each data point based on the predicted strength of the corresponding formula. Figure~\ref{fig:strength_env_profile} shows the strength spectrum of the newly generated concrete formulas plotted in the environmental impact space for each concrete curing time group. These plots could serve as a visualization tool for concrete designers to quickly select newly generated formulas that meet the design requirements.

\subsection{Strength-Conditioned Progression Generation}

\begin{figure}
    \centering 
  \includegraphics[width=1.8in]{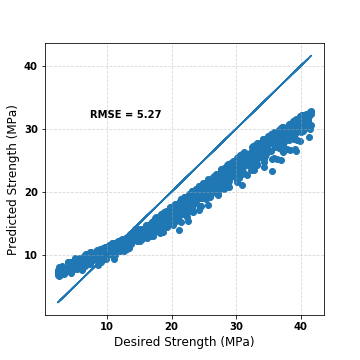}
  \includegraphics[width=1.8in]{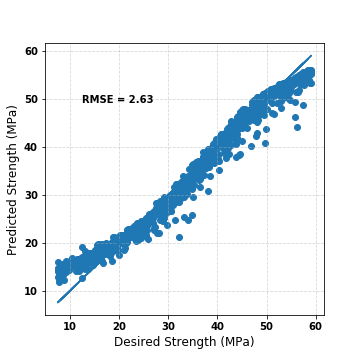}
 \includegraphics[width=1.8in]{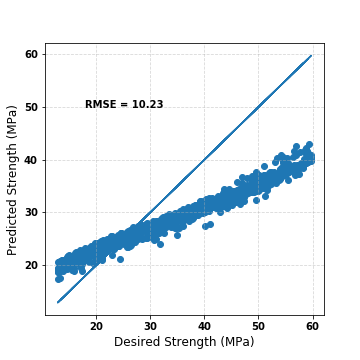} \\

(a) \qquad\qquad\qquad\qquad\qquad\qquad\qquad (b) \qquad\qquad\qquad\qquad\qquad\qquad\qquad (c) \\ 

\includegraphics[width=1.8in]{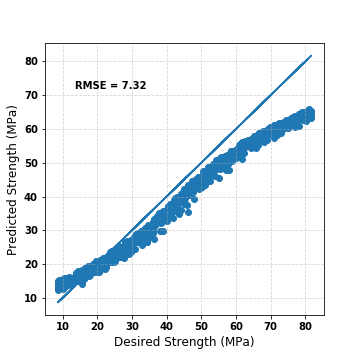}
\includegraphics[width=1.8in]{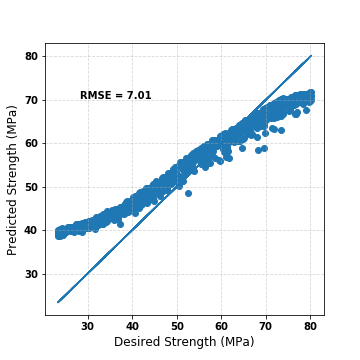}
\includegraphics[width=1.8in]{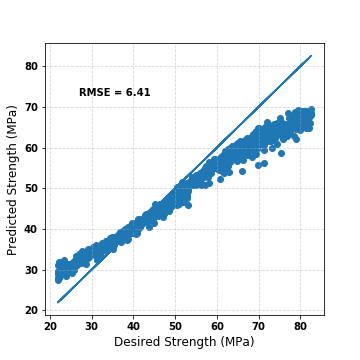}

(d) \qquad\qquad\qquad\qquad\qquad\qquad\qquad (e) \qquad\qquad\qquad\qquad\qquad\qquad\qquad (f)

\caption{Strength-conditioned progression for different concrete curing times. (a) $\leq$ 3 days. (b) 7 days. (c) 14 days. (d) 28 days. (e) 56 days. (f) $\geq$90 days.}
\label{fig:strength_progression}
\end{figure}

Attribute-conditioned generative progression has been considered by \cite{SohnLY2015}. In the context of face image generation there, one of the attribute dimension values such as gender, facial expression, or hair color is modified by interpolating between the minimum and maximum attribute value, i.e.\ $x = [x_\alpha, x_{rest}]$, where $x_\alpha = (1-\alpha)\cdot x_{min} + \alpha\cdot x_{max}$. Indeed, one can visualize that the attribute of generated images change progressively with the change in conditioning attribute values. 

To further demonstrate our concrete formulation generator can produce concrete designs with desired properties, we perform similar experiments. For the purpose of illustration, we limit our conditioning variables to strength and curing time of the concrete. We again generate 10,000 samples for each curing time group, by uniformly sampling from $[x_{min}, x_{max}]$. Figure~\ref{fig:strength_progression} shows how well the predicted strength of generated formulations match with the desired strength given as conditioning variable during generation. The performance varies across different curing time groups. The RMSE is computed to evaluate the performance quantitatively. The better performing model should have the contour of the scattered dots to cover the diagonal line. The result shows that the generator seems to work the best for concrete curing time of 7 days.

\section{Laboratory Experiments}

Five concrete formulations were generated using the CVAE approach, aiming to minimize environmental impacts under given compressive strength targets.  Since the nature of superplasticizer has changed since the training dataset was developed, human adjustment of superplasticizer proportion was made to improve rheology (and address this drift from historical data); in particular, all superplasticizer quantities were set to $1/4$ of their AI-generated values.  Proportions for the concrete mixes that were made for laboratory experiments are listed in Table~\ref{tab:mix_composition}.

\begin{table}
    \centering
\begin{tabular}{|p{0.6cm}|p{1cm}|p{1.7cm}|p{1cm}|p{1.7cm}|p{1.9cm}|p{1.4cm}|p{1.6cm}|}
\hline
mix & cement [kg/m$^3$] & blast furnace slag [kg/m$^3$] & fly ash [kg/m$^3$] & water (used) [kg/m$^3$] & superplasticizer (used) [kg/m$^3$] & coarse aggregate [kg/m$^3$] & fine aggregate [kg/m$^3$] \\ \hline
1 & 131.46 & 201.21 & 119.67 & 180.70 & 1.1 & 950.72 & 780.48 \\ \hline
2 & 128.59 & 197.46 & 124.24 & 184.31 & 1.0 & 954.48 & 787.47 \\ \hline
3 & 134.89 & 182.74 & 113/78 & 179.43 & 0.9 & 953.22 & 785.28 \\ \hline
4 & 132.25 & 184.37 & 119.74 & 181.03 & 1.8 & 954.10 & 786.55 \\ \hline
5 & 129.02 & 210.60 & 122.80 & 184.63 & 1.0 & 953.50 & 780.11 \\ \hline
\end{tabular}
\caption{Concrete mixes that underwent laboratory testing.}
    \label{tab:mix_composition}
\end{table}

To manufacture the concrete mixes, ordinary Portland cement (OPC), blast furnace slag, fly ash, and potable water were used for making the cement paste. Limestone and natural sand were used as coarse and fine aggregate, respectively. Both these aggregates meet the particle size distribution requirements from the ASTM standards. An industrial pan mixer was used for the mixing process.  

\subsection{Mixing Procedure}
The properties of fresh concrete were investigated with concrete that was mixed in accordance with the ASTM C192 standard. First, the pan was wet with water, and the excess water was wiped out with a towel. Next, coarse aggregate, fine aggregate, and half of the water were mixed for 30 seconds. After 30 seconds, the mixer was stopped, and the cementitious materials were added to the pan. The mixer was restarted, and the remaining water was added carefully to the pan. The concrete was mixed for 3 minutes, rested for 1 minute in which the superplasticizer was added, and then mixed for an additional 2 minutes, as specified by the ASTM standards. 

\subsection{Slump Test}
A slump test for fresh concrete was then conducted. The test was conducted in accordance with the ASTM C143 standard. First, the slump cone was placed on the mat and secured to the floor by stepping on the foothold of the cone. The cone was filled with fresh concrete up to 1/3 of its total volume and rodded 25 times with the 5/8-inch (15.8 mm) tamping rod. These steps were repeated two more times until the slump cone was entirely filled with concrete. The top of the slump cone was leveled with the trowel. Finally, the slump cone was carefully pulled directly upward in a period of 3 to 7 seconds, and the slump value was measured using a scale.

\subsection{Preparation of Concrete Cylinders}
Cylindrical specimens were prepared with a diameter of 4 inches (101.6 mm) and a length of 8 inches (203.2 mm) to be used for the compressive strength tests, according to ASTM C31. A total of 6 cylindrical specimens were made for each mix, three to be tested for compressive strength after 7 days, and the other three to be tested after 28 days. To prepare the cylinders, the cylindrical molds were first oiled to assist in demolding. Any excess oil was then wiped out with a towel to ensure there was only a thin layer of form oil. The first half of the cylinder was filled with concrete, rodded 25 times with a small tamping rod, then the outside of the mold was tapped 10 to 15 times to eliminate the air voids. The second half of the mold was then filled, tamped, and tapped in the same exact manner as the first half. The top of the mold was then finished with the trowel, and the cylinders were covered with a plastic sheet and left for 24 hours to harden. After hardening, the specimens were demolded and left at a temperature of 68°F (20°C) in the moist cure room.

\subsection{Compressive Strength Tests}
A Forney machine was used to conduct compressive strength tests on the concrete cylinders after 7 and 28 days of curing and in accordance with the ASTM C39 standard. To perform the test, the cylindrical specimen was first capped with sulfur-based capping compound on the top and bottom and placed on the center of the loading platen. The load was applied at a constant rate of 27,000 lb/min (12,240 kg/min) until the specimen failed, and the ultimate load was recorded. Also, the fracture pattern was noted.  

\begin{figure}
\centering
\includegraphics[width=5in]{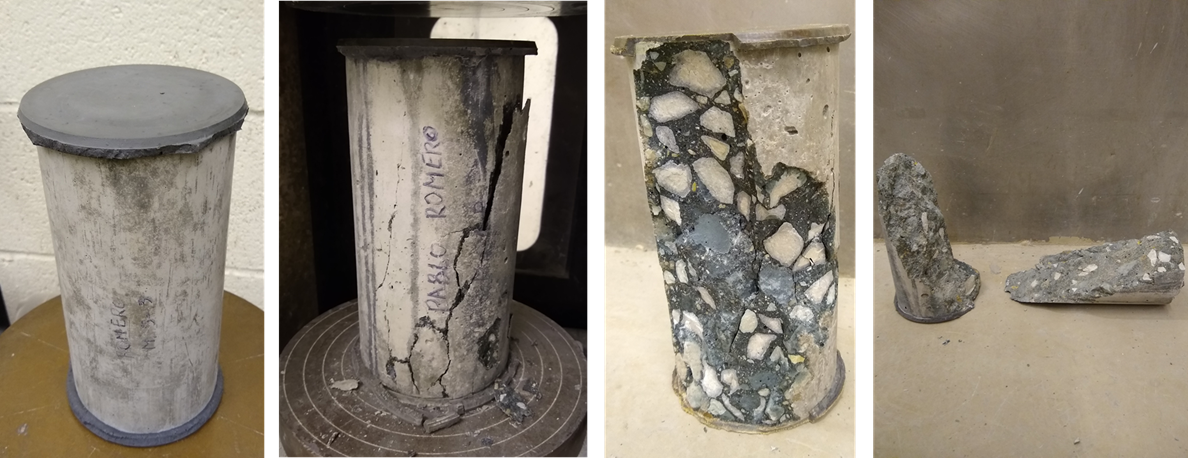}
\caption{Photographs of some concrete cylinders made according to novel formulations, undergoing laboratory testing for compressive strength.}
\label{fig:cylinders}
\end{figure}

\begin{figure}
\centering
\includegraphics[width=3.5in]{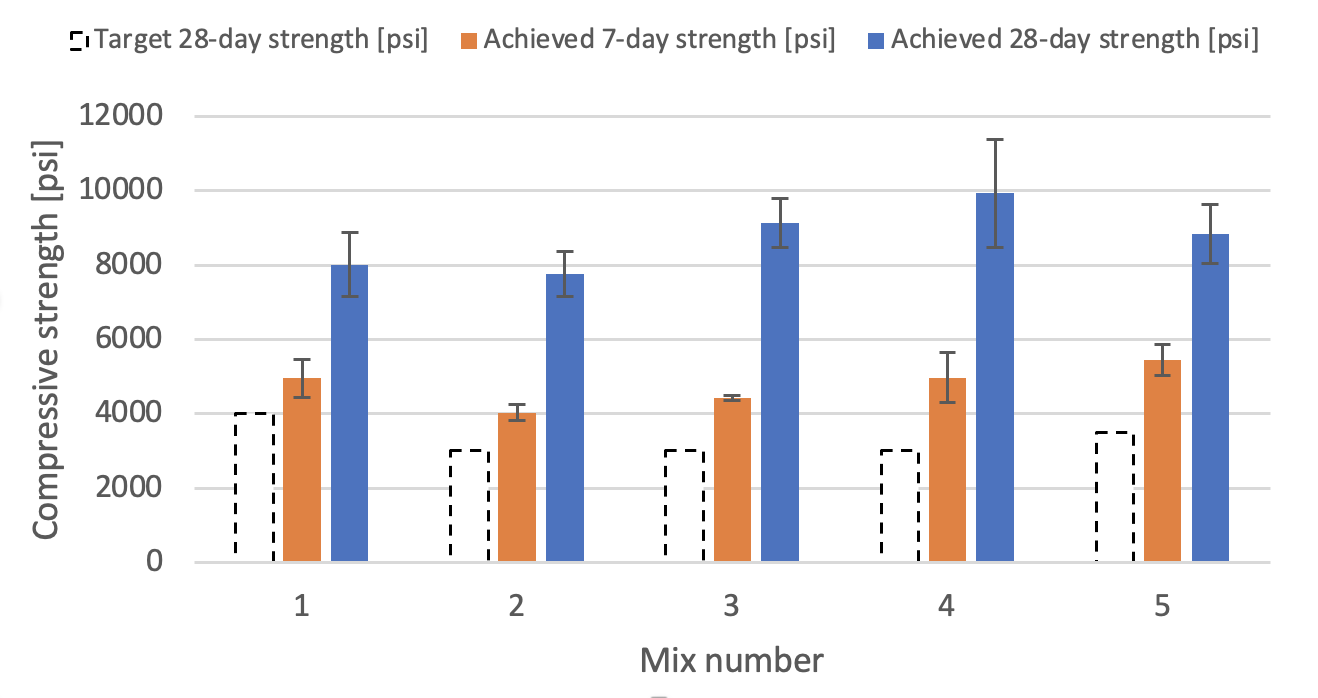}
\caption{Experimental compressive strength results for five AI-generated concrete mixes.  Error bars correspond to three cylinders each.}
\label{fig:compressivestrength}
\end{figure}

Figure~\ref{fig:cylinders} shows some of the concrete cylinders, whereas Figure~\ref{fig:compressivestrength} shows compressive testing results.  These results are further quantified in Table~\ref{tab:mix_labresults}. Note that all five tested mixes exceed the target 28-day strength already at 7 days and that the 28-day measured strength was double or more than the target. Table~\ref{tab:mix_labresults} shows results from the slump test.

\begin{table}
    \centering
\begin{tabular}{|p{0.6cm}|p{2cm}|p{2cm}|p{2cm}|p{1cm}|p{2cm}|p{1.8cm}|p{1cm}|} \hline
mix & 28 day strength target [psi] & 7 day strength [psi] & 28 day strength  [psi] & slump [in] & global warming potential [kg CO$_2$ eq./$m^3$] & acidification potential [kg SO$_2$ eq./$m^3$] & batch water [m$^3$] \\ \hline
1 & 4000 & 4949 ± 507 & 8013 ± 871 & 5 1/2 & 154.111 & 0.534 & 0.181 \\ \hline
2 & 3000 & 4025 ± 217 & 7764 ± 610 & 3 1/4 & 152.284 & 0.530 & 0.183 \\ \hline
3 & 3000 & 4431 ± 66 & 9136 ± 656 & 6 3/4 & 157.294 & 0.547 & 0.180 \\ \hline
4 & 3000 & 4967 ± 673 & 9938 ± 1453 & 5 3/4 & 155.157 & 0.549 & 0.167 \\ \hline
5 & 3500 & 5443 ± 417 & 8836 ± 801 & 3 3/4 & 152.149 & 0.524 & 0.169 \\ \hline
\end{tabular}
\caption{Laboratory testing and lifecycle analysis results.}
    \label{tab:mix_labresults}
\end{table}

\subsection{Global Warming Potential}
Table~\ref{tab:mix_labresults} also shows the lifecycle analysis-based computations for the GWP, AP, and CBW environmental metrics.  Table~\ref{tab:compare} explicitly demonstrates that these formulations roughly halve the global warming potential as compared to the average of similar 28-day compressive strength formulations. To make this comparison, we used the achieved strength numbers from Table~\ref{tab:mix_labresults} against similar achieved numbers from the UCI Machine Learning Repository \cite{Yeh1998}. 

Further comparison could be made against the industry standard given in \cite{NRMCA2019}, which gives regional performance benchmarks. Importantly, this provides an external comparison beyond the training dataset that was used in our work. For the Great Lakes region that the field test associated with this paper belongs to, this report gives the following benchmarks, against which our numbers compare favorably: 
\begin{itemize}
    \item $3000$ psi: $281.33$ kg CO$_2$/m$^3$
    \item $4000$ psi: $334.87$ kg CO$_2$/m$^3$
\end{itemize}

\begin{table}
    \centering
    \begin{tabular}{|p{2in}|p{1cm}|p{1cm}|p{1cm}|p{1cm}|p{1cm}|} \hline
    mix  & 1 & 2 & 3 & 4 & 5 \\ \hline
    Estimated GWP (kg CO$_2$ eq./$m^3$) & 154.11 & 152.28 & 157.29 & 155.16 & 152.15 \\  \hline
    Average of industry standard of similar 28-day compressive strength (kg CO$_2$ eq./$m^3$) & 282.36 & 280.31 & 318.75 & 302.45 & 279.78 \\ \hline
    \end{tabular}
    \caption{Comparing global warming potential of AI-generated mixes to average of industry standard mixes of similar 28-day compressive strength.}
    \label{tab:compare}
\end{table}

\section{Field Tests}

\subsection{Adjustments for Logistical and Environmental Factors}

The lab-tested formulations -- in particular Mix 1 with 28-day compressive strength target of 4000 psi -- were provided to concrete supplier Ozinga Ready Mix of Chicago, IL. Further lab testing was done, and changes were made to the original formula to take two factors into account:
\begin{enumerate}
    \item Logistical considerations, for example the availability of certain materials.
    \item Potentially cold weather conditions at the Meta Data Center construction site, which will affect curing speed.
\end{enumerate}
Mix 1 contained significant amounts of cementitious material, with fly ash and slag being use as cement replacement material. However, availability of fly ash and slag is limited in certain locations and times. Further, keeping the relative proportion of cement as compared to the cement replacement materiel was taken into consideration, given the cold weather conditions. As a result of these, the formula was adjusted, as given in Table~\ref{tab:pours}, for two separate field tests. 

\subsection{Deployment in Field Tests}

After internal testing and adjustments by Ozinga, the formulations  were deployed in two separate pours of slab-on-grade foundations. A slab-on-grade is a type of shallow foundation in which a concrete slab rests directly on the ground below it. A slab-on-grade foundation usually consists of a thin layer of concrete across the entire area of the foundation with thickened footings at the edges. 

\begin{figure}
    \centering
    \includegraphics[width=2.5in]{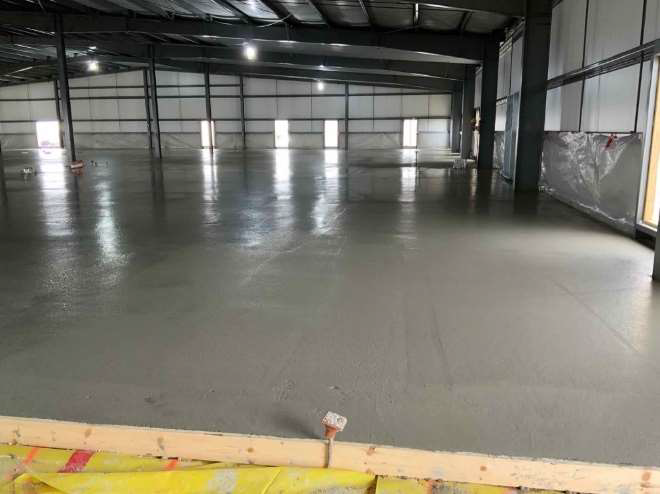} \ \includegraphics[width=2.5in]{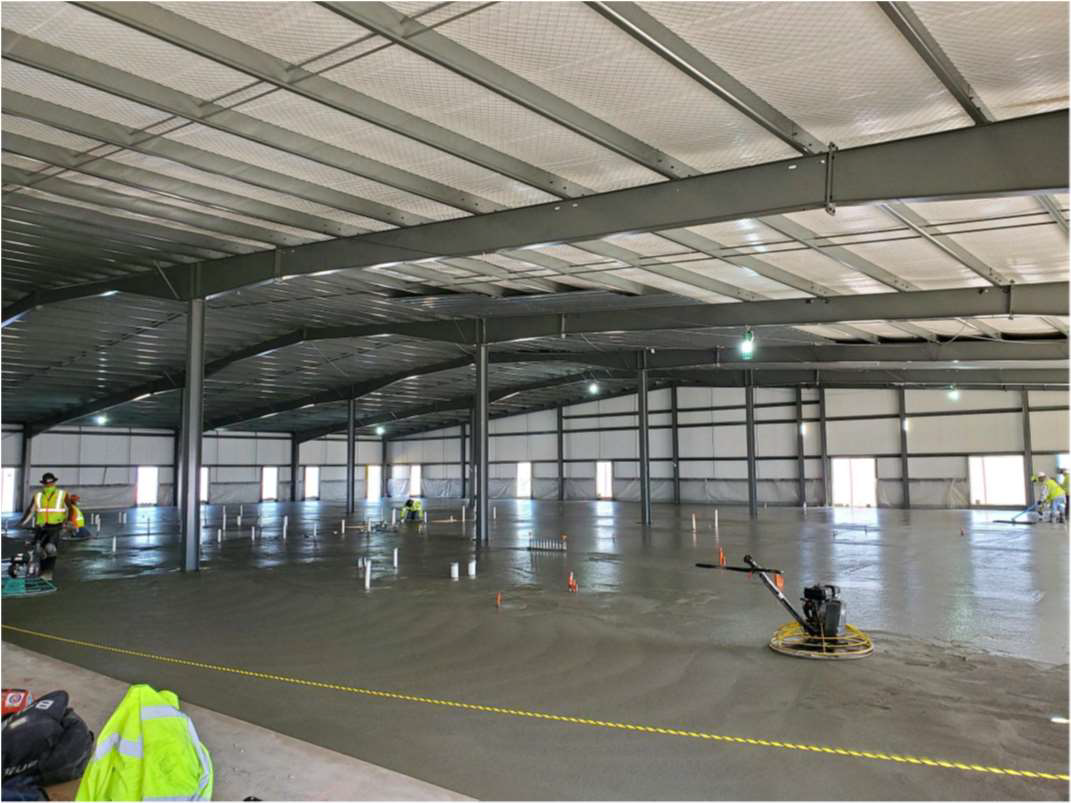} \\ (a) \qquad\qquad\qquad\qquad\qquad\qquad\qquad\qquad\qquad (b)
    \caption{Pictures of Pour 1 on day of deployment. (a) first half. (b) second half.}
    \label{fig:pour1_im}
\end{figure}

\begin{figure}
    \centering
    \includegraphics[width=2.5in]{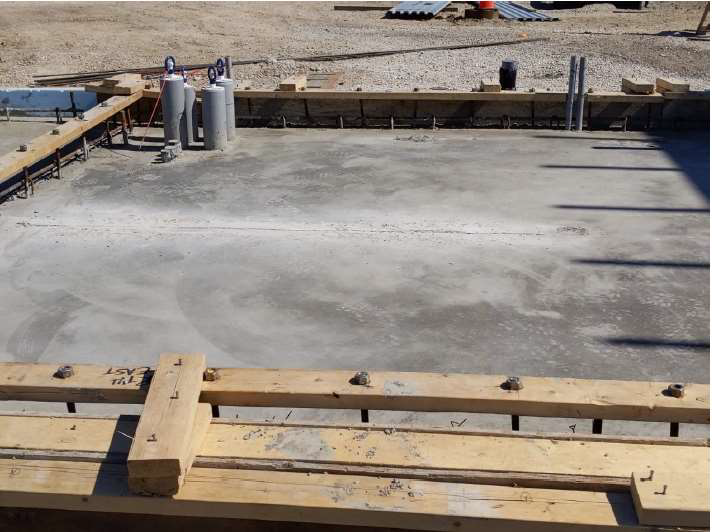} \ \includegraphics[width=1.4in]{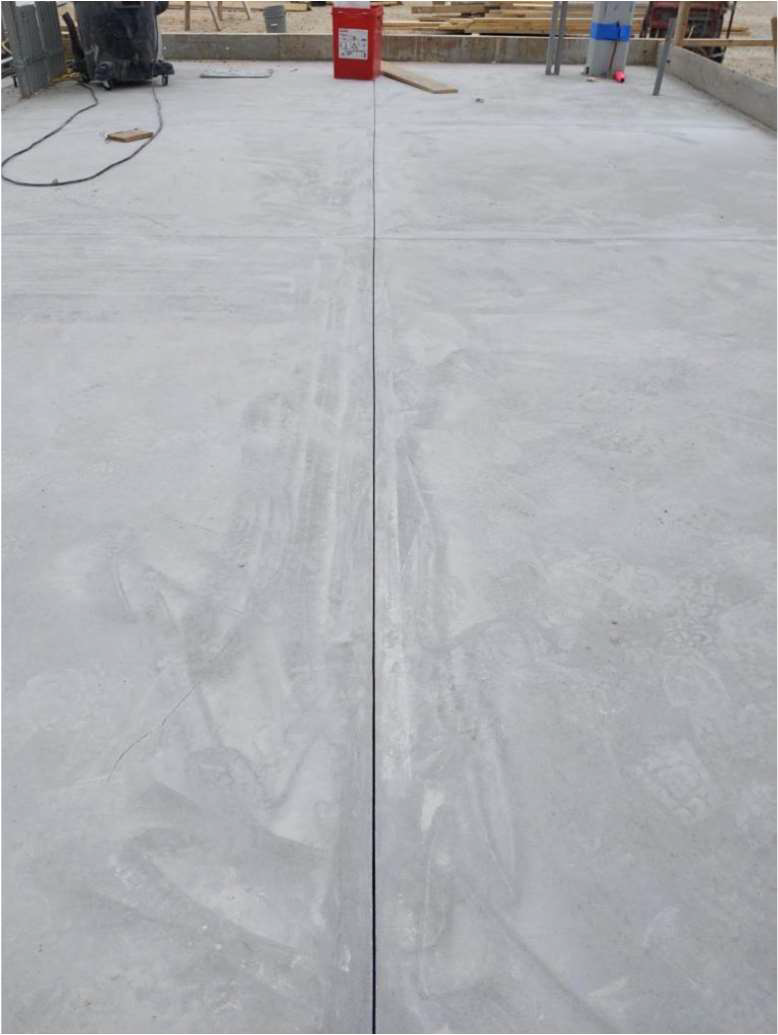} \\ (a) \qquad\qquad\qquad\qquad\qquad\qquad\qquad\qquad\qquad (b)
    \caption{Pictures of Pour 2. (a) the foundation. (b) the saw cut made into the foundation.}
    \label{fig:pour2_im}
\end{figure}

In Pour 1, the concrete was used for an office building used for the construction crew. This slab-on-grade foundation of size 40,000 square feet, see Figure~\ref{fig:pour1_im}, was done in March 2021 during cold days, with temperatures in (42°F--52°F). In Pour 2, the concrete was used for a guard house.  This slab-on-grade foundation of size 450 square feet, see Figure~\ref{fig:pour2_im}, was done in June 2021 during hot days, with temperatures in (80°F--98°F). 

\begin{table}
    \centering

    \begin{tabular}{|l|l|l|l|} \hline
    & Pour 1 & Pour 2 & Mix 1 \\ \hline
    Ingredient & Amount per cubic yard & Amount per cubic yard & Amount per cubic yard \\ \hline
    Cement (ASTM C 150 Type I/II) & 215 lb & 215 lb & 211.6 lb \\ \hline
    Fly Ash (ASTM C 618 Class C) & 100 lb & 100 lb & 339.2 lb \\ \hline
    Slag (ASTM C 989) & 180 lb & 180 lb & 201.7 lb \\ \hline
    Water – Potable & 25.3 gal & 24.0 gal & 36.5 gal \\ \hline
    Coarse Aggregate (ASTM C 33) & 1930 lb & 1920 lb & 1602.5 lb \\ \hline
    Fine Aggregate (ASTM C 33) & 1500 lb & 1485 lb & 1315.5 lb \\ \hline
    Water-Cementitious Material Ratio & 0.43 & 0.40 & 0.40 \\ \hline
    \end{tabular}
    
    \caption{Compositions of two field-deployed concrete mixes. Note that the mixes also included CarbonCure, Accelerator (ASTM C 494 Type C), and Water Reducer (ASTM C 494 Type F) whose admixture dosage rates varied based on concrete temperature, ambient temperature, haul time, etc.}
    \label{tab:pours}
\end{table}

The quantitative results from these field deployments are given in Tables \ref{tab:pour1_results}--\ref{tab:pour2_results}.  As can be observed, all specifications were met, despite the variability in weather and other environmental conditions.  The only concern raised by the contractor was the slow curing time of Pour 1 in cold weather, in missing the unspecified, but still desirable, 5-day strength at 75\% of 28-day target. Early strength performance at 1, 3, or 5-day may be needed specifically for slab-on-grade applications to allow further activities such as saw cutting and other finishing.  The qualitative results, such as visual aesthetics, were also acceptable.

\begin{table}
    \centering
    \begin{tabular}{|l|l|l|l|} \hline
    Performance Metric & Specification & Result & Note\\ \hline
    3-day strength & --- & 1936 psi & no spec, but better than expected \\ \hline
    5-day strength & --- & 2650 psi & no spec, but less than expected \\ \hline
    7-day strength & 3000 psi & 3510 psi & passed spec \\ \hline
    28-day strength & 4000 psi & 6970 psi & passed spec \\ \hline
    Slump & 5--7 in. & 6.42 in. & passed spec \\ \hline
    Air content & 0--3\% & 1.3\% & passed spec \\ \hline
    Shrinkage & < 0.032\% at 21 days (ASTM C157) & 0.022\% & passed spec \\ \hline
    \end{tabular}
    \caption{Quantitative results from first half of Pour 1: compressive strengths are average of seven cylinders; slump and air content are averages of three samples.}
    \label{tab:pour1_results}
\end{table}

\begin{table}
    \centering
    \begin{tabular}{|l|l|l|l|} \hline
    Performance Metric & Specification & Result & Note \\ \hline
    3-day strength & --- & 1620 psi & no spec, but better than expected \\ \hline
    5-day strength & --- & 2667 psi & no spec, but less than expected \\ \hline
    7-day strength & 3000 psi & 3540 psi & passed spec \\ \hline
    28-day strength & 4000 psi & 7592 psi & passed spec \\ \hline
    Slump & 5--7 in. & 6.25 in. & passed spec \\ \hline
    Air content & 0--3\% & 1.6\% & passed spec \\ \hline
    Shrinkage & < 0.032\% at 21 days (ASTM C157) & 0.022\% & passed spec \\ \hline
    \end{tabular}
    \caption{Quantitative results from second half of Pour 1: compressive strengths are average of seven cylinders; slump and air content are averages of three samples.}
    \label{tab:pour1b_results}
\end{table}

\begin{table}
    \centering
    \begin{tabular}{|l|l|l|l|} \hline
    Performance Metric & Specification & Result & Note \\ \hline
    3-day strength & --- & 1960 psi & no spec, but better than expected \\ \hline
    7-day strength & --- & 2940 psi & no spec, but as expected \\ \hline
    28-day strength & 4000 psi & 6260 psi & passed spec \\ \hline
    Slump & 5--7 in. & 5.50 in. & passed spec \\ \hline
    Air content & 0--3\% & 2.2\% & passed spec \\ \hline
    Shrinkage & < 0.032\% at 21 days (ASTM C157) & 0.033\% & passed spec (okay within 15\%) \\ \hline
    \end{tabular}
    \caption{Quantitative results from Pour 2: compressive strengths are average of seven cylinders; slump and air content are averages of three samples.}
    \label{tab:pour2_results}
\end{table}

\section{Conclusion}
We have demonstrated end-to-end accelerated design and deployment of low-carbon concrete, based on a CVAE model that discovers new concrete formulations with reduced environmental impact, including embodied carbon, without sacrificing strength. The resulting formulations were then further optimized based on environmental and logistical constraints. 

However, there is still room for improving our model and broader approach. Our data contains both continuous and categorical values, but CVAE may not be the best for capturing such mixed categorical and continuous data. The VAE-ROC model proposed by \cite{SuhC2016} is said to be better at handling mixed data. We hope by modifying the CVAE model in line with specifics of the VAE-ROC, the generator would synthesize more realistic concrete designs and achieve better performance in attribute-conditioned generation. Faster curing of concrete formulations in cold weather is also desired as an additional property. This includes 1, 3, and 5-day curing performance which are important for mission-critical applications such as slab-on-grade. 

Finally, as we have noted, the availability of cement replacement materials such as fly ash and slag vary with location and time. Therefore there is an opportunity to directly optimize for such logistical considerations. Further, there is need to identify and discover novel materials that could be used in addition to, or in replacement of ,such cementitious materials. 

\begin{acks}
Discussions with  Jeremy Gregory, Randolph Kirchain, and Joana Maria are very much appreciated.

This work was supported in part by an unrestricted gift from Meta. XG, HY, and LRV were also supported in part by the IBM-Illinois Center for Cognitive Computing Systems Research (C3SR), a research collaboration as part of the IBM AI Horizons Network.
\end{acks}

\bibliographystyle{ACM-Reference-Format}
\bibliography{concrete}

\end{document}